\documentclass[runningheads]{llncs}

\usepackage{eccv}

\usepackage{eccvabbrv}

\usepackage{graphicx}
\usepackage{booktabs}
\usepackage{bm}
\usepackage{enumitem}
\usepackage{tikz}
\usepackage{pifont}
\usepackage{algorithm}
\usepackage{algpseudocode}
\usepackage{float}
\usepackage{pifont}
\usepackage{multirow}

\usepackage{array} 
\usepackage{caption} 
\usepackage{subcaption}
\usepackage[accsupp]{axessibility}  
\newcommand{\repeatthanks}{\textsuperscript{\thefootnote}}

\usepackage{hyperref}

\usepackage{orcidlink}

\begin{document}

\title{\texttt{Gen-Swarms}: Adapting Deep Generative Models to Swarms of Drones} 

\titlerunning{\texttt{Gen-Swarms}}

\author{Carlos Plou\thanks{equal contribution; $^{\dagger}$ corresponding authors}$^{\dagger}$\inst{1}\orcidlink{0009-0004-7535-9955} \and
Pablo Pueyo\repeatthanks$^{\dagger}$\inst{1}\orcidlink{0000-0001-6482-009X}
\and
Ruben Martinez-Cantin \inst{1}\orcidlink{0000-0002-6741-844X}
\and
Mac Schwager\inst{2}\orcidlink{0000-0002-7871-3663} 
\and
Ana C. Murillo \inst{1}\orcidlink{0000-0002-7580-9037
}
\and
Eduardo Montijano \inst{1}\orcidlink{0000-0002-5176-3767}}

\authorrunning{C. Plou et al.}

\institute{DIIS-i3A, University of Zaragoza, Spain \\
\email{{\{c.plou,ppueyor,rmcantin,acm,emonti\}@unizar.es}}\and
Dept. of Aeronautics and Astronautics, Stanford University, USA\\
\email{schwager@stanford.edu}}

\maketitle

\begin{abstract}
\texttt{Gen-Swarms} is an innovative method that leverages and combines the capabilities of deep generative models with reactive navigation algorithms to automate the creation of drone shows. Recent advancements in deep generative models, particularly diffusion models, have demonstrated remarkable effectiveness in generating high-quality 2D images. Building on this success, various works have extended diffusion models to 3D point cloud generation. 
In contrast, alternative generative models such as flow matching have been proposed, offering a simple and intuitive transition from noise to meaningful outputs. However, the application of flow matching models to 3D point cloud generation remains largely unexplored.
\texttt{Gen-Swarms} adapts these models to automatically generate drone shows. Existing 3D point cloud generative models create point trajectories which are impractical for drone swarms. In contrast, our method not only generates accurate 3D shapes but also guides the swarm motion, producing smooth trajectories and accounting for potential collisions through a reactive navigation algorithm incorporated into the sampling process.
For example, when given a text category like \textit{Airplane}, \texttt{Gen-Swarms} can rapidly and continuously generate numerous variations of 3D airplane shapes. Our experiments demonstrate that this approach is particularly well-suited for drone shows, providing feasible trajectories, creating representative final shapes, and significantly enhancing the overall performance of drone show generation. Code has been released to the community \footnote{Homepage located at: \url{https://cplou99.github.io/Gen-Swarms/}}.

\end{abstract}

\section{Introduction}

The development of foundational models in deep learning has paved the way for significant advancements across various domains. Foundational models, characterized by their ability to learn from vast amounts of data and generalize across multiple tasks, have involved innovative applications. Among these, deep generative models have emerged as powerful methods for data generation and manipulation.

Deep generative models aim to map a simple distribution, such as a Gaussian, to a more complex target distribution. Many of these methods train a model to learn how to iteratively refine noise into a final coherent output that lies within the target distribution. Among these approaches, diffusion models have gained popularity due to their success in generating 2D images conditioned on given text queries \cite{Ho2020}. Thanks to their success, diffusion models have been explored for more complex domains such as videos, molecules, and 3D point clouds \cite{chai2023stablevideo, hoogeboom2022equivariant, mo2024dit}. Recently, flow matching \cite{lipmanflow}, an alternative to diffusion models, which combines the best features of diffusion models and normalizing flows, has gained attention \cite{klein2024equivariant, pooladian2023multisample}. This new approach has shown excellent performance in generating simple 2D images, indicating its potential for broader applications \cite{mehta2024matcha, yim2023fast}.

\begin{figure}[!t]
    \centering
    \includegraphics[width=\linewidth]{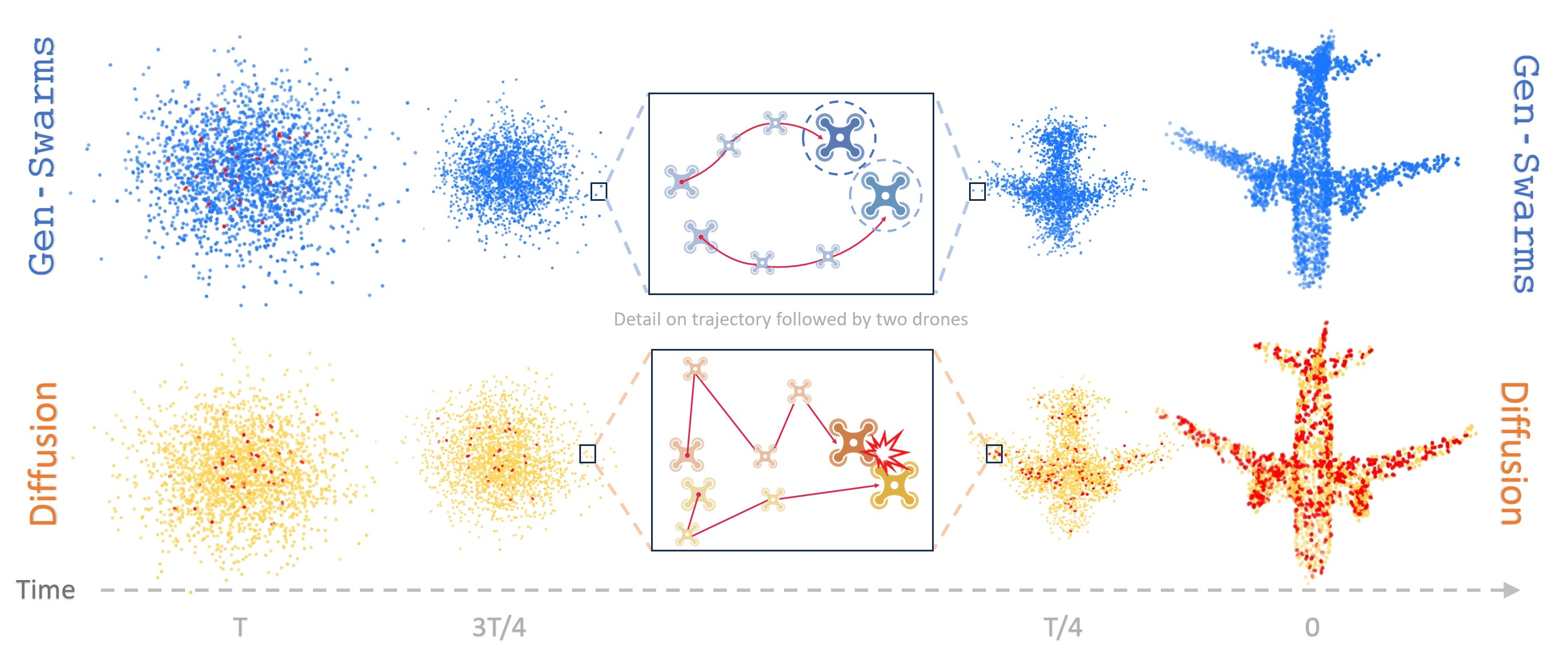}
    \caption{Illustration of \texttt{Gen-Swarms}, a novel \textbf{3D point cloud generative model adapted to handle motion constraints in physical swarms}. Unlike the nearest current approach (\emph{Diffusion} \cite{luo2021diffusion}), \texttt{Gen-Swarms} produces \textbf{smooth} trajectories (zoomed in the middle) free of \textbf{collisions} (red points indicates collisions).}
    \label{fig:header}
\end{figure}

The use of drones in collaborative art (usually called ``drone shows'') has become a trend for both the robotics and entertainment communities \cite{waibel2017drone}. In these shows, drones act as pixels to create complex and dynamic shapes in the sky, captivating audiences of even major events such as the Olympic ceremonies or sports shows. 
Despite their impressive results and success, these displays often require manual design and intervention, limiting their scalability, variety, and affordability. This highlights the need for more autonomous and adaptable solutions in drone show generation.

Applying generative models to drone shows offers a promising solution, as they can learn to generate realistic and complex 3D shapes while providing all the intermediate steps to reach the goal. The closest application of current generative models to drone swarms is 3D point cloud generation \cite{mo2024dit, vahdat2022lion, liu2023meshdiffusion}, as each point can be treated as a drone. However, a direct application of these methods is impractical. This is because the trajectories of the points generated by these models are unfeasible for guiding swarm motion due to two main factors: (1) lack of smoothness, and (2) continuous collisions.  In other words, unlike current 3D point clouds generative model applications that primarily focus on the quality of the generated output and efficiency of the model, our emphasis is on the intermediate steps leading to the final output.

This paper introduces \texttt{Gen-Swarms} (\cref{fig:header}), a novel framework that leverages the strengths of 3D point clouds generative models, flow matching and reactive navigation algorithms to automatically generate drone shows. This framework adapts to the motion constraints of physical swarms, which are not addressed by existing applications. Specifically, by inputting a class category (e.g., ``airplane''), \texttt{Gen-Swarms} can rapidly and continuously generate safe and smooth trajectories, guiding the drones to form accurate and diverse 3D shapes that represent the specified category.

Our experimental results indicate that our method is particularly effective for this application, providing safer and smoother trajectories for the drones than baselines, representing a significant advancement in the field of collaborative artistic robotics. Our main contributions are:

\begin{itemize}[label=$\bullet$]
    \item Introduction of \texttt{Gen-Swarms}, a novel collaborative robotics framework that leverages deep generative models and reactive navigation algorithms to create drone shows based on a desired shape category. This framework addresses challenges in swarm robotics, such as dynamic constraints and collision avoidance.
    \item The first application of flow matching algorithms to 3D point cloud generation. We incorporate an encoder to obtain a latent space representation of the 3D point cloud distribution, which conditions the posterior sampling process.
\end{itemize}
\label{sec:intro}

\section{Related Work}

\noindent
\textbf{3D Point Clouds Generation.} 3D point cloud generation aims to create novel and accurate three-dimensional objects from scratch. This field has seen significant advancements through the application of diverse approaches such as normalizing flows \cite{yang2019pointflow, kim2020softflow, klokov2020discrete}, variational autoencoders \cite{kim2021setvae, yang2018foldingnet, gadelha2018multiresolution}, and generative adversarial networks \cite{valsesia2018learning, achlioptas2018learning, shu20193d}. Current state-of-the-art approaches rely on  Denoising Diffusion Probabilistic Models (DDPMs) \cite{Ho2020}. These models utilize a forward noising process that gradually adds Gaussian noise to raw data and trained a reverse process that inverts the forward process. Notable works in this domain include DPM \cite{luo2021diffusion}, the first DDPM approach to learn the reverse diffusion process for point clouds as a Markov chain conditioned on shape latent. Other works are based on the point-voxel representation \cite{zhou20213d} and graph structure \cite{liu2023meshdiffusion}  of 3D shapes. Differently, GET3D \cite{gao2022get3d} is conditioned to two latent codes (a 3D Signed Distance Function and a texture field) and LION \cite{vahdat2022lion} proposes two hierarchical DDPMs in global latent and latent point spaces.
Lastly, DiT-3D \cite{mo2024dit} proposes a novel Diffusion Transformer that can directly operate the denoising process on voxelized point clouds. Based on DDPMs, another work \cite{wu2023fast} proposed a novel method for fast point cloud generation using straight velocity flows. Instead, we focus on obtaining feasible trajectories for drone swarms.

\noindent
\textbf{Flow Matching.} Flow matching (FM) \cite{lipmanflow} is a relatively new strategy for making Continuous Normalizing Flows trainable, which has rapidly gained popularity in the deep learning community \cite{song2024equivariant, mehta2024matcha}. FM aligns data distributions through an Optimal Transport conditional path, enabling the learning of the flow. Initial tests on ImageNet demonstrate that FM consistently outperforms alternative diffusion-based methods in both likelihood and sample quality, but its main advantage is the simple and intuitive paths it provides \cite{lipmanflow}. Building upon this foundational work, several subsequent studies have explored the application of flow matching to various domains. For instance, flow-based generative models have been employed in video prediction \cite{davtyan2023efficient},  molecular design \cite{song2024equivariant, yim2023fast}, and speech generation \cite{liugenerative, mehta2024matcha}, demonstrating their versatility and robustness. In this work, we propose flow matching algorithm conditioned on shape latent to 3D point cloud generation task, and we adapt it to drone swarms constraints.

\noindent
  \textbf{Collaborative artistic robotics.} In the realm of multirobot systems, research has focused on creating artistic formations using multiple robots or drones. Early work in this area involved optimal control techniques to arrange robots into predefined 2D patterns, demonstrating the feasibility of coordinated multi-robot formations \cite{alonso2011multi, alonso2012image, hauri2013multi}. More recent studies have expanded this work to multidrone formations, enabling complex visual displays and coordinated performances. These advancements have leveraged sophisticated algorithms and control mechanisms to achieve precise and dynamic aerial formations, often used in entertainment and advertising \cite{nar2022optimal, waibel2017drone, kim2016realization}. Typically, these works receive the desired pattern as an input, guiding the robots or drones to form specific shapes or images.

In contrast, the objective of \texttt{Gen-Swarms} is to generate shapes from a single word, pushing the boundaries of generative modeling in robotic formations. A recent approach, ClipSwarm, optimizes a formation of robots using foundational models to form a desired shape based on 2D images defined by a contour. However, this method is currently limited to 2D representations \cite{pueyo2024clipswarm}. By integrating advanced generative models, \texttt{Gen-Swarms} aims to enhance the creative and autonomous capabilities of multirobot systems, facilitating the formation of accurate 3D shapes from minimal input.
\label{sec:related}

\section{Background}

\subsection{Generative modeling}
Let us assume we have data samples $\mathbf{x}^{(1)}_0, \ldots, \mathbf{x}^{(n)}_0$ drawn from a distribution of interest, whose density $p_{0}(\mathbf{x})$ is unknown. Our goal is to utilize these samples to learn a probabilistic model $p_{\theta}(\mathbf{x})$ that approximates this distribution $p_{0}(\mathbf{x})$. Specifically, we aim to efficiently generate new samples from $ p_{\theta}(\mathbf{x})$ that are approximately distributed according to the target distribution. This task is known as \textbf{generative modeling}.

To make sampling possible, it is usually assumed a simple prior distribution $p_T(\mathbf{x})$ (e.g., normal distribution), and, then, the problem becomes finding a neural transport map between both distributions $p_T(\mathbf{x})$ and $p_{0}(\mathbf{x})$. Within this framework, several strategies are prominent, particularly Denoising Diffusion Probabilistic Models (DDPMs) and Continuous Normalizing Flows (CNFs).

\subsection{DDPMs and CNFs}

\textbf{Denoising Diffusion Probabilistic models (DDPMs)} \cite{Ho2020, Song2020} suggested an iterative noising forward process that gradually adds gaussian noise to raw data 
\begin{equation}
\mathbf{x}_t = \sqrt{1-\beta_{t}}\mathbf{x}_{t-1} + \sqrt{\beta_t}\epsilon, \;\; \epsilon\sim \mathcal{N}(\mathbf{0}, \mathbf{I}), \;\;\; t\in \{1, \ldots, T\},    
\end{equation}
where $\{\beta_t\}_{t=1}^{T}$ are hyperparameters.

For the reverse process, diffusion models are trained to learn a denoising network for inverting
forward process corruption as $p_{\theta}(\mathbf{x}_{t-1} |\mathbf{x}_t) = \mathcal{N}(\mu_{\theta}(\mathbf{x}_t, t), \sigma^{2}_{t} \mathbf{I})$. Once $p_{\theta}(\mathbf{x}_{t-1} |\mathbf{x}_t)$ is trained, new data points can be generated by progressively sampling $\mathbf{x}_{t-1} \sim p_{\theta} (\mathbf{x}_{t-1}|\mathbf{x}_t)$ with initialization $x_{T} \sim
\mathcal{N} (\mathbf{0}, \mathbf{I})$. However, as demonstrated in \cref{sec:experiments}, this iterative sampling process results in very noisy paths unfeasible for actual drones.

Alternatively, \textbf{Continuous Normalizing Flows (CNFs)} \cite{Chen2018} target a probability density function continuous path $p_t, \; t \in [0, T]$ between the simple prior \( p_T \) and our distribution of interest $p_0$. Due to this continuity, $T$ is typically set to $1$. This transition path may be generated by a parameterized vector field $v_{\theta}(\mathbf{x}, t)$, which defines through ordinary differential equation (ODE), the flow $\phi(\mathbf{x}, t)$,

\begin{equation}
\frac{d}{dt} \phi(\mathbf{x}, t) = v_{\theta}(\phi(\mathbf{x}, t), t); \quad \phi (\mathbf{x}, T) = \mathbf{x},
\end{equation}
 where $ p_t $ can be derived with the change of variables 
 \( p_t = p_T (\phi^{-1}(\mathbf{x}, t)) \det \left[ \frac{\partial \phi_t^{-1}}{\partial \mathbf{x}} \right] \). Despite it ensures a deterministic sampling, this strategy requires expensive numerical ODE simulations at training time.

\subsection{Flow matching for generative modeling}\label{sec:FM}

\textbf{Flow matching (FM)} is a strategy for CNFs training. FM directly formulates a regression objective w.r.t. the parameterized vector field $v_{\theta}(\mathbf{x}, t) $
of the form%
\begin{equation}
L_{FM} (\theta) = \mathbb{E}_{t \sim \mathcal{U}[0,T], x_{t} \sim p_t (x)} \left| \left| v_{\theta}(\mathbf{x}_{t}, t)  - v_{*} (\mathbf{x}_{t}, t) \right| \right|^2,
\end{equation}
where $ v_{*} (\mathbf{x}_{t}, t)$ is a target vector field inducing a probability path $p_t$
interpolating from $p_T$ to $p_0$. Hence, the key is to approximate $v_{*} (\mathbf{x}_t, t)$.
To achieve this goal, \textbf{Conditional Flow Matching (CFM)} \cite{lipmanflow} suggests to  
adopt the Optimal Transport conditional path,
\begin{equation}
    p_t (\mathbf{x}|\mathbf{x}_0) := \mathcal{N} \left( \mathbf{x} | t' \mathbf{x}_0, \sigma_t \mathbf{I} \right),   \;\; \sigma_t  = 1 - (1 - \sigma_{\min}) t', \;\; t'=(T-t)/T,
\end{equation}
with a sufficiently small \( \sigma_{\min} \)  such that \( p_0 (\mathbf{x}|\mathbf{x}_0) \) is centered around \( \mathbf{x}_0 \). It is also observed that $p_T (\mathbf{x}|\mathbf{x}_0)=\mathcal{N} \left( \mathbf{x} | \mathbf{0}, \mathbf{I} \right)$. In other words, similar to DDPMs, as time goes by, the points gradually diffuse
into a chaotic set of points. This parameterization yields tractable target vector fields,
\begin{equation}
    v_{*} (\mathbf{x}_{t}, t) \simeq v_{*} (\mathbf{x}| \mathbf{x}_{0}, t) = \frac{\mathbf{x}_{0} - (1 - \sigma_{\min}) \mathbf{x}}{1 - (1 - \sigma_{\min}) t'}.
\end{equation}
 In this scenario, CFM training objective can be simplified to,%
\begin{equation}
L_{CFM} (\theta) = \mathbb{E}_{t \sim \mathcal{U}[0,T],  \mathbf{x}_0 \sim p_0 (\mathbf{x})} \left| \left| v_{\theta} (\mathbf{x}_t, t) - \left( \mathbf{x}_0 - (1 - \sigma_{\min}) \mathbf{x}_{T} \right) \right| \right|^2,
\end{equation}
where $\mathbf{x}_{t} \sim p_t (\mathbf{x}|\mathbf{x}_0)$ and $\mathbf{x}_{T} \sim \mathcal{N} \left( \mathbf{x} | \mathbf{0}, \mathbf{I} \right)$. Once the model is trained, we may generate new samples solving the next ODE integration.
\begin{equation}
    \mathbf{x}_0 = \mathbf{x}_T + \int_{T}^{0} v_{\theta}(\mathbf{x}_t, t)dt,  \;\; \mathbf{x}_{T} \sim \mathcal{N} \left( \mathbf{x} | \mathbf{0}, \mathbf{I} \right).
\end{equation}
\label{sec:background}

\section{\texttt{Gen-Swarms}}

\texttt{Gen-Swarms} is a generative model that can simultaneously provide the final point cloud and guide the swarm motion to reach it. Consequently, our main challenge is to ensure smooth and collisions-free trajectories. In this section, we first  formulate the method and, then, describe the training and sampling implementation.

\subsection{Method}
Let $\mathbf{X}=\{\mathbf{x}^{i}\}_{i=1}^{N} \in \mathbb{R}^{N \times 3}$ be a point cloud consisting of $N$ 3D-points, where each point $\mathbf{x}^i$ represents a drone. We adopt the Optimal Transport conditional path $p_t (\mathbf{X}|\mathbf{X}_0), \; t \in [0, T]$ suggested by CFM technique (\cref{sec:FM}), 
\begin{equation}
    p_t (\mathbf{X}|\mathbf{X}_0) := \mathcal{N} \left( \mathbf{X} | t' \mathbf{X}_0, \sigma_t \mathbf{I} \right),   \;\; \sigma_t  = 1 - (1 - \sigma_{\min}) t', \;\; t'=(T-t)/T.
\end{equation}

\noindent From here, the three main components of our method (which is graphically described in \cref{fig:oursmethod}) are:
\begin{figure}[!t]
    \centering
\includegraphics[width=\linewidth]{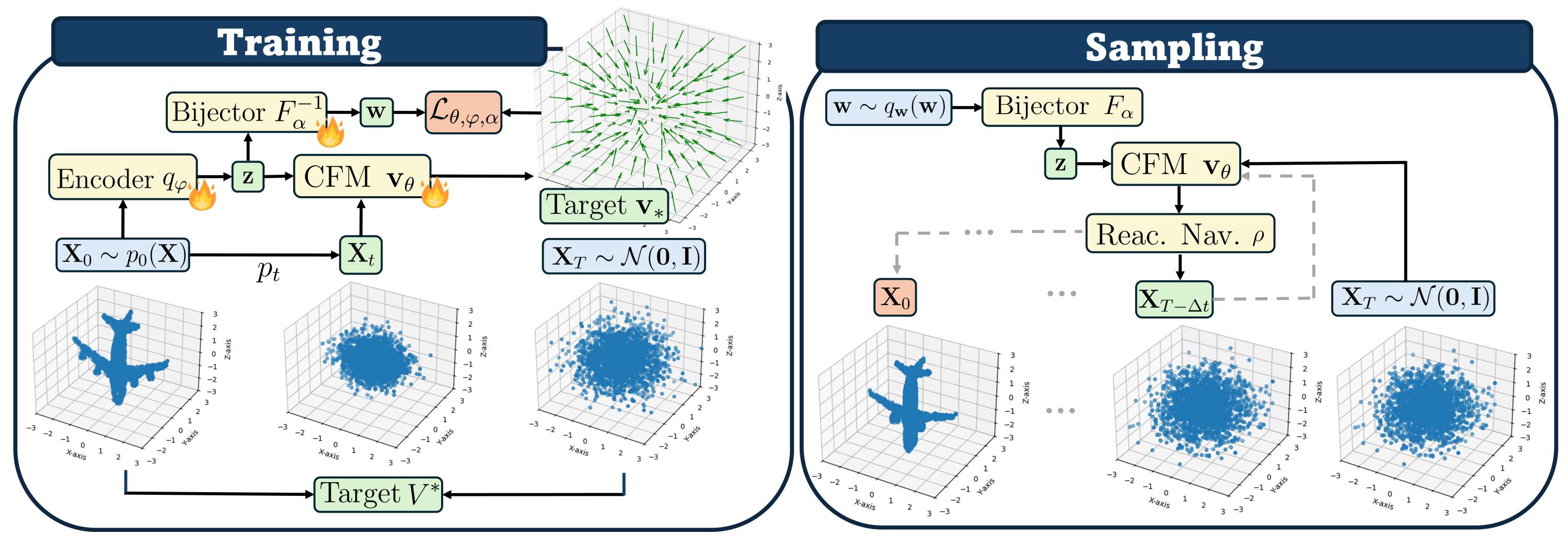}
    \caption{Illustration of \texttt{Gen-Swarms}. \textbf{On the left}, we show one iteration of the \textbf{training algorithm}. Starting (blue) from the sampled point clouds, we derive intermediate calculations (green), to end (red) computing the loss $\mathcal{L}_{\theta, \varphi, \alpha}$. This loss is then used to perform backpropagation through our three neural networks (yellow). \textbf{On the right}, the \textbf{sampling algorithm} generates safe and smooth trajectories from a random initial point cloud towards an accurate 3D shape guided by the shape latent $\mathbf{z}$.}
    \label{fig:oursmethod}
\end{figure}

\paragraph{\textbf{CFM network-.}} This probability path defines a target vector field that will determine the desired velocity of each drone $i$ depending on its initial position $\mathbf{x}^{i}_0$ and the time $t$, 
\begin{equation}
        \mathbf{v}_{*} (\mathbf{x}^{i}| \mathbf{x}^{i}_{0}, t) = \frac{\mathbf{x}^{i}_{0} - (1 - \sigma_{\min}) \mathbf{x}^{i}}{1 - (1 - \sigma_{\min}) t'}.
\end{equation}
This target vector field will be estimated from a point cloud $\mathbf{X}_t$ by $\mathbf{v}_{\theta}(\mathbf{x}^{i}_{t}, t, \mathbf{z})$ implemented by a neural network parameterized by $\theta$. Specifically, we opt for a Gated Contextual Network formed by several linear layers that combine the features $\mathbf{X}_t$ with the shape latent $\mathbf{z}$ and the temporal context $t$ through a gating mechanism (sigmoid function).

\paragraph{\textbf{Encoder-.}} As it is noticed, we incorporate the shape latent $\mathbf{z} \in \mathbb{R}^d$ as input in order to easily guide the transition towards the desired shape. This d-dimensional latent space is simultaneously learned by an encoder $q_\varphi(\mathbf{z} | \mathbf{X}_0)$ parameterized by $\varphi$ whose loss compares the output distribution $q_\varphi(\mathbf{z} | \mathbf{X}_0)$ with a prior distribution $q(\mathbf{z})$ using the KL-divergence $D_{KL}\left(q_\varphi(\mathbf{z} | \mathbf{X}_0) |  q(\mathbf{z}) \right)$. In this context, we adopt PointNet \cite{qi2017pointnet} as the encoder $q_{\varphi}(\mathbf{z} | \mathbf{X}_0)$.

\paragraph{\textbf{Bijector-.}} The most common choice of the prior is a simple fixed distribution as $q(\mathbf{z}):=\mathcal{N}(\mathbf{0}, \mathbf{I})$. However, an interesting approach is to give more flexibility to this prior distribution and make it trainable \cite{luo2021diffusion}. Specifically, we opt for a trainable bijector $F_{\alpha}$ formed by several affine coupling layers that maps an isotropic Gaussian $q_{\mathbf{w}}(\mathbf{w})$ to a complex distribution $q(\mathbf{z})$. This bijection mapping enables us to compute the complex distribution $q(\mathbf{z})$  by the change-of-variable,  
\begin{equation}
 q(\mathbf{z}) = q_{\mathbf{w}}\left (\mathbf{w} \right) \bigg|  \frac{\partial F_{\alpha}}{\partial \mathbf{w}} \bigg|^{-1}, \;\; \mathbf{w}= F^{-1}_{\alpha}(\mathbf{z}).
\end{equation}
In other words, it enables us to train the neural network in the opposite direction that will be used during inference. 

Therefore, the training loss $\mathcal{L}_{\theta, \varphi, \alpha}$, which integrates the training processes of the three modules $\mathbf{v}_\theta, q_\varphi, F_{\alpha}$ presented before, can be summarized as follows, 
\begin{equation}\label{eq:loss}
\mathcal{L}_{\theta, \varphi, \alpha} =  \mathbb{E}_{t,  \mathbf{X}_0}  \left[ \sum_{i=1}^{N}\left\| \mathbf{v}_{*} (\mathbf{x}^{i}_{t}| \mathbf{x}^{i}_{0}, t) - \mathbf{v}_\theta \left(\mathbf{x}^{i}_t, t , \mathbf{z}\right) \right\|^2 + D_{KL}\left(q_\varphi(\mathbf{z} | \mathbf{X}_0) | q( \mathbf{z}) \right) \right].
\end{equation}

Once the method is trained, we could generate new data samples $\mathbf{X}_{0}=\{\mathbf{x}^{i}_{0}\}_{i=1}^{M}$ as follows,

\begin{equation}\label{Eq:Int}
    \mathbf{x}^{i}_0 = \mathbf{x}^{i}_T + \int_{T}^{0} \mathbf{v}_{\theta}(\mathbf{x}^{i}_t, t, F_{\alpha}(\mathbf{w}))dt,  \;\;\;  \mathbf{X}_{T}, \mathbf{w} \sim \mathcal{N} \left(\mathbf{0}, \mathbf{I} \right).
\end{equation}

The integral of \cref{Eq:Int} could be approximated by a numerical integration method. As we will see later, the resulting trajectories are smooth, successfully addressing one of our two major challenges. However, additional implementations are needed to tackle the collision avoidance challenge. Thus, we suggest to discretize the integral of \cref{Eq:Int}, computing $\mathbf{v}_{\theta}(\mathbf{x}^{i}_t, t, F_{\alpha}(\mathbf{w}))$ at $\Delta t$ rate and incorporate a reactive navigation method that at each step $t$ refines the predicted vector field to avoid collisions. This reactive navigation method $\rho$ solves an optimization problem by minimally adjusting the velocities $\mathbf{V}_{\theta}^{t} = [\mathbf{v}_{\theta}(\mathbf{x}^{1}_{t}, t, F_{\alpha}(\mathbf{w}))^{T}, \ldots, \mathbf{v}_{\theta}(\mathbf{x}^{M}_{t}, t, F_{\alpha}(\mathbf{w}))^{T}]$ of the drones located at $\mathbf{X}_t$ to ensure a minimum security distance $\kappa$. In mathematical terms,%
\begin{equation}\label{Eq:Ours}
    \mathbf{X}_{t-\Delta t} = \mathbf{X}_{t} + \rho( \mathbf{V}^{t}_{\theta}, \mathbf{X}_{t}, \kappa),  \;\;\;\; \forall t \in \{T, T-\Delta t, \ldots, \Delta t\}.
\end{equation}%

\noindent Interestingly, these slight modifications, which could be interpreted as noise for the CFM network, do not prevent reaching an accurate final point cloud.

\subsection{Implementation}

\subsubsection{Training.}
To simplify and enhance the efficiency of the training process, we adopt a sampling approach \cite{Ho2020}. Instead of computing the expectation by integrating over the time space $[0, T]$ in \cref{eq:loss}, we randomly select one term from the summation to optimize during each training step. Specifically, this training algorithm is summarized in \cref{alg:training}.

\begin{algorithm}[!tb]
\caption{Training}
\begin{algorithmic}[1]
\State $\sigma_{\min} = 10^{-4}$
\Repeat
    \State $\mathbf{X}_0 \sim p_{0}(\mathbf{X})$ \Comment{Point cloud sample}
    \State $\mathbf{z} \sim q_\varphi(\mathbf{z} | \mathbf{X}_0)$ \Comment{Shape latent}
\State $t \sim \mathcal{U}(0, T)$ \Comment{Time step sampling}
\State $t' = (T-t)/T$
\State $\pmb{\epsilon} \sim \mathcal{N}(\mathbf{0}, \mathbf{I})$ \Comment{Noise}
\State $\mathbf{X}_t = (1 - (1 - \sigma_{\min})t') \pmb{\epsilon} + t' \mathbf{X}_0$ \Comment{Point cloud defined by the OT path}
\State $\mathbf{V}_* = \mathbf{X}_0 - (1 - \sigma_{\min}) \pmb{\epsilon}$ \Comment{Optimal flow}
\State $ \mathbf{V}_\theta^{t}=[\mathbf{v}_{\theta}(\mathbf{x}^{1}_t, t, F_{\alpha}(\mathbf{w}))^{T}, \ldots, \mathbf{v}_{\theta}(\mathbf{x}^{N}_t, t, F_{\alpha}(\mathbf{w}))^{T}]$
\State $\mathcal{L}_{\theta, \varphi, \alpha} = \left\| \mathbf{V}_* - \mathbf{V}_\theta^{t}\right\|^2 + D_{KL}\left(q_\varphi(\mathbf{z} | \mathbf{X}_0) | q( \mathbf{z}) \right)
    $  \Comment{Loss }
\State Take gradient descent step on
    $
    \nabla_{\theta, \varphi, \alpha} \mathcal{L}_{\theta, \varphi, \alpha}
    $ \Comment{Backpropagation}
\Until{converged}
\end{algorithmic}
\label{alg:training}
\end{algorithm}

\subsubsection{Sampling.}
We can generate a safe and smooth trajectory towards a new 3D point cloud following \cref{Eq:Ours}. A key aspect is that the number of points  $M$ during sampling can differ from the number of points $N$ during training, as the CFM network acts as a closed-loop controller for each drone in the swarm, with the shape latent $\mathbf{z}$ serving as the reference. The pseudocode for this step is detailed in Algorithm~\ref{alg:sampling}.

\begin{algorithm}[!tb]
\caption{Sampling}
\begin{algorithmic}[1]
\State $w \sim \mathcal{N}(\mathbf{0}, \mathbf{I})$ 
\State $\mathbf{z} \sim F_{\alpha}(w)$ \Comment{Shape latent}
\State $\mathbf{X}_T \sim \mathcal{N}(\mathbf{0}, \mathbf{I})$ \Comment{Point cloud sample}
\For{$t \in \{T, T-\Delta t, \ldots, \Delta t\}$}
\State $\mathbf{V}_{\theta}^{t} = [\mathbf{v}_{\theta}(\mathbf{x}^{1}_{t}, t, F_{\alpha}(\mathbf{w}))^{T}, \ldots, \mathbf{v}_{\theta}(\mathbf{x}^{M}_{t}, t, F_{\alpha}(\mathbf{w}))^{T}]$ \Comment{Predicted flow}
\State $\mathbf{X}_{t-\Delta t} =\mathbf{X}_{t} + \rho(\mathbf{V}_{\theta}^{t}, \mathbf{X}_{t},  \kappa)$ \Comment{Next point cloud}
\EndFor
\end{algorithmic}
\label{alg:sampling}
\end{algorithm}
\label{sec:approaches}

\section{Experiments}
\label{sec:experiments}

\subsection{Dataset}
 In line with previous works on 3D point cloud generation \cite{zhou20213d, mo2024dit}, we utilize the ShapeNet dataset \cite{chang2015shapenet}, which comprises 51,127 shapes across 55 categories. Similar to previous methods, we focus on the \emph{Airplane} category for reporting metrics and quantitative results. However, we also provide qualitative results for other categories within the ShapeNet dataset. During training, we use all $N=2,048$ points of each point cloud and normalize each point cloud to an isotropic Gaussian distribution. For inference, while we have the flexibility to choose the number of points in the generated final point cloud, we set it to $M=2,048$ points for shape completion to maintain consistency. Afterward, we scale the point clouds to a realistic drone-show size, corresponding to a $200 \times 200 \times 200$ meter space, suitable for a fleet of drones.

\subsection{Experimental Setup}\label{Sec:setup}
\paragraph{\textbf{Method details-.}}
Regarding architecture details, our Gated Contextual Network consists of $6$ layers, while the bijector $F_{\alpha}$ comprises 14 coupling layers. These neural networks, along with the PointNet encoder with embedding size $d=256$, are trained using the Adam optimizer with a linear scheduler for the learning rate. The experiments were conducted on an Intel Core™ i7-12700K processor with 20 cores and an NVIDIA GeForce RTX 3090 GPU. Furthermore, focusing on method hyperparameters, we set $T = 1$ and vary the $\Delta t$ rate according to the experiment to ensure a fair comparison. Furthermore, we set a minimum security distance between points of $\kappa = 2$ meters in our final scale, resulting in $\kappa = 0.06$ meters on the training scale. This distance is based on the assumption that drones are 0.5 meters in size (0.25m of radius) with a security distance of 1.5 meters between drones. Lastly, we choose ORCA \cite{ORCA} as our reactive navigation method $\rho$.

\paragraph{\textbf{Evaluation metrics-.}} According to the challenges discussed, we analyze metrics based on four different criteria: (1) Quality of the final shape generated, (2) Smoothness of the trajectory, (3) Collisions, and, (4) Energy consumption. 
\begin{itemize}
    \item \textbf{Quality of the 3D shapes}:  We draw from 3D point cloud generative models~\cite{luo2021diffusion, mo2024dit} and employ Chamfer Distance (CD) as distance measure. We use CD to calculate the Coefficient of Variation (COV) as a diversity measure and the minimum matching distance (MMD) to measure the fidelity of the generated samples.
    \item \textbf{Collisions}: We define collision as a violation of the minimum security distance $\kappa$ between two drones. Then, we compute the average percentage of drones violating this distance $\kappa$ along the whole trajectory (TRAJ) and also only for the final 3D shape (FIN).
    \item\textbf{Smoothness}: We analyze point (drone) velocities along the paths. Specifically, we calculate variations in the magnitude (ACC), in $m/s^2$, the third derivative of the positions (JERK), in $ m/s^3$, and the velocity direction (DIR), in radians. Lower values of these metrics indicate fewer variations in the trajectories, resulting in smoother and more realistic drone movements~\cite{alcantara2021optimal}.
    \item \textbf{Energy}: to assess the energy consumption, we compute the mean distance traveled (DIST) by the drones along the entire trajectory.
\end{itemize}

\paragraph{\textbf{Baselines-.}} To the best of our knowledge, there is no relevant prior work that specifically uses generative models for drone shows. Therefore, we implement three baselines to facilitate a meaningful comparison with our method, \texttt{Gen-Swarms}.

\begin{itemize}
\item \emph{Diffusion}: We developed a Diffusion-based approach leveraging DPM \cite{luo2021diffusion}. This method utilizes the same architecture as \texttt{Gen-Swarms} but employs diffusion algorithms for both training and sampling. This baseline allows us to evaluate the impact of using the CFM algorithm in our method.
\item \emph{CFM}: To ensure a fair comparison with the \emph{Diffusion} approach, we evaluate the performance of the \emph{CFM} algorithm alone. For this baseline, we train as in \texttt{Gen-Swarms} but perform sampling from \eqref{Eq:Int} instead of \eqref{Eq:Ours}.
\item \emph{CFM+ORCA}: This third baseline is designed to assess the performance of \texttt{Gen-Swarms} sampling. This approach takes the final point cloud generated by \emph{CFM} and uses directly the ORCA reactive navigation method to generate all trajectories from the random initial point cloud towards the goal.
\end{itemize}

\subsection{Quantitative results}
For each method, we generate 607 samples, matching the test set size to obtain accurate quality metrics for 3D shapes. Additionally, to ensure a fair comparison, we perform sampling over 100 steps ($\Delta t = 0.01s.$) for all methods. We compare the performance of \texttt{Gen-Swarms} against the baselines in the \emph{Airplane} category using the metrics outlined in \cref{Sec:setup}. The results, presented in a real scale of $200\times200\times200$ meters, are summarized in Table~\ref{table:exp_results}. As observed, \emph{Diffusion} leads in quality but falls short in collision avoidance, smoothness and energy consumption. \emph{Conditional Matching Flow} improves the smoothness and provides more direct, lower-energy trajectories, but still suffers from collisions. When leveraging ORCA to generate safe trajectories towards the final 3D point clouds generated by CFM, it fails to reach these positions within the same number of steps, and does not form the final desired shape, as graphically visualized in the fourth column of Fig. \ref{fig:final_shapes_collisions}. In contrast, \texttt{Gen-Swarms} delivers balanced performance, excelling in collision avoidance (collisions are only present in the initial random point cloud, which is why the TRAJ metric is not zero), energy efficiency, and maintaining high quality while ensuring smooth trajectories. This is because \texttt{Gen-Swarms} uses CFM as a closed-loop controller with ORCA as an additional reactive navigation layer inside the loop.

\begin{table}[!t]
\begin{center}
\caption{\textbf{Metrics comparison: \texttt{Gen-Swarms} vs. baselines}. Quality of reconstructions, percentage of collisions, smoothness, and energy consumption in the \emph{Airplane} category. Best results in bold, second best underlined.}
\begin{tabular}{ c | c  c | c c | c  c  c | c  }
\toprule
Model & \multicolumn{2}{c|}{QUALITY - CD} & 
\multicolumn{2}{c|}{COLLS (\%, $\downarrow$)} & \multicolumn{3}{c|}{SMOOTH ($\downarrow$)} & \multicolumn{1}{c}{EN. ($\downarrow$)} \\
\hline 
 & COV ($\uparrow$) & MMD ($\downarrow$) & TRAJ & FIN  & ACC & JERK & DIR & DIST  \\
\hline 
\emph{Diffusion} & \textbf{0.49} & \textbf{29.81} & 11.31 & 48.98  &  0.97 & 7.05 & 12.19 & 495.25 \\
\emph{CFM} & 0.45 & 32.24  & 15.38  & 42.87 &  0.54 & 0.01 & \textbf{0.003}  & \underline{44.07}  \\
\emph{CFM+ORCA} & 0.01 & 2836.74 & \underline{0.05} & \underline{0.02}  & \textbf{0.008} & \underline{0.005} & \underline{0.05} & 46.45   \\
\hline 
\texttt{Gen-Swarms} & \underline{0.46} & \underline{31.87} & \textbf{0.04} & \textbf{0}  & \underline{0.014} & \textbf{0.004} & 0.54 & \textbf{44.04}   \\
\bottomrule
\end{tabular}
\label{table:exp_results}
\end{center}
\end{table}

\begin{figure}[!t]
\includegraphics[width=\linewidth]{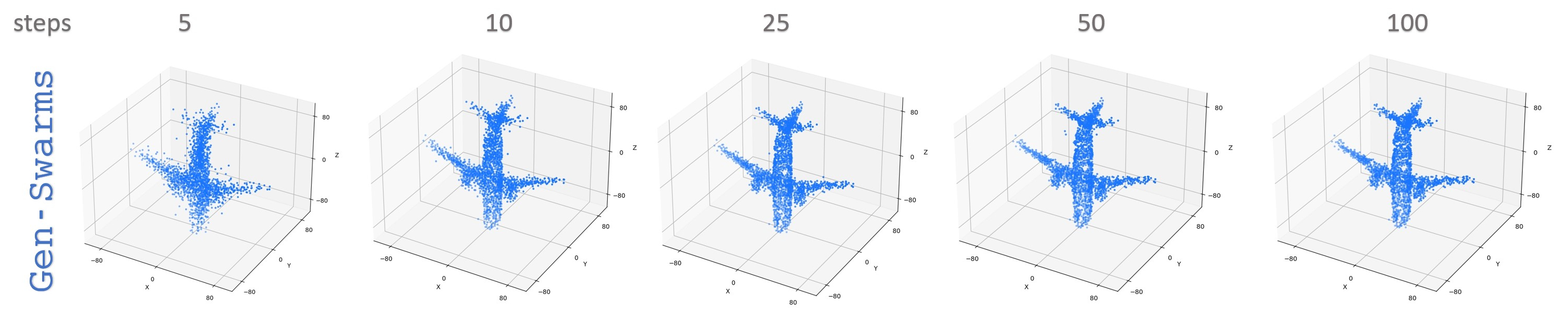}
    \caption{\textbf{Impact of $\Delta t$ in the \texttt{Gen-Swarms} quality reconstruction}. The final 3D shapes generated by \texttt{Gen-Swarms} for the different number of steps highlight the convergence at $\Delta t=0.04s.$ (25 steps).}
    \label{fig:steps}
\end{figure}

Next, we focus on the influence of the 
 sampling rate $\Delta t$ (which determines the number of steps) on the performance of \texttt{Gen-Swarms}, as shown in Table~\ref{table:steps}. This indicates that while the quality of the generated 3D shapes stabilizes at around 25 steps (also observed in \cref{fig:steps}), further increasing the number of steps enhances the overall safety and smoothness of the generated drone trajectories. Conversely, a higher number of steps makes it more challenging to run our method online, as the reactive navigation method could become a computational and temporal bottleneck in the pipeline.  Therefore, depending on the device used, the optimal number of steps should be determined to balance this trade-off.

\begin{table}[!t]
\begin{center}
\caption{\textbf{Impact of $\Delta t$ in the \texttt{Gen-Swarms} sampling.} Quality of reconstructions, percentage of collisions, smoothness, and energy consumption metrics for different $\Delta t$ values in the \emph{Airplane} category. Best results in bold, second best underlined.}
\begin{tabular}{ c | c  c | c c | c  c  c | c  }
\toprule
\texttt{Gen-Swarms} & \multicolumn{2}{c|}{QUALITY - CD} & 
\multicolumn{2}{c|}{COLLS (\%, $\downarrow$)} & \multicolumn{3}{c|}{SMOOTH ($\downarrow$)} & \multicolumn{1}{c}{EN. ($\downarrow$)} \\
\hline 
 Steps ($\Delta t$) & COV ($\uparrow$) & MMD ($\downarrow$) & TRAJ & FIN  & ACC & JERK & DIR & DIST  \\
\hline 
5 (0.2) & 0.29 & 43.09 & 119 & 395 & 5.40 & 5.32 & 0.64 & \textbf{43.21} \\
10 (0.1)  & 0.42 & 34.15 & 46.16 & 315.26 &  1.39 & 0.97 &  0.58 & 44.05   \\
25 (0.04) & \textbf{0.46} & \underline{31.85} & 3.85 &  11.56 & 0.23 & 0.11 & \underline{0.55} &  47.74  \\
50 (0.02) & \underline{0.45} & \textbf{31.81} & \underline{1.68} & \underline{0.58} & \underline{0.058} & \underline{0.02} & \underline{0.55} & 44.04   \\
100 (0.01) & \textbf{0.46} & 31.87 & \textbf{0.83} & \textbf{0}  & \textbf{0.014} & \textbf{0.004} & \textbf{0.54} & \underline{44}  \\
\bottomrule
\end{tabular}
\label{table:steps}
\end{center}
\end{table}

\subsection{Qualitative results}
Complementing the quantitative results, we provide some qualitative results.

\begin{figure}[!tb]
    \centering
    \includegraphics[width=\linewidth]{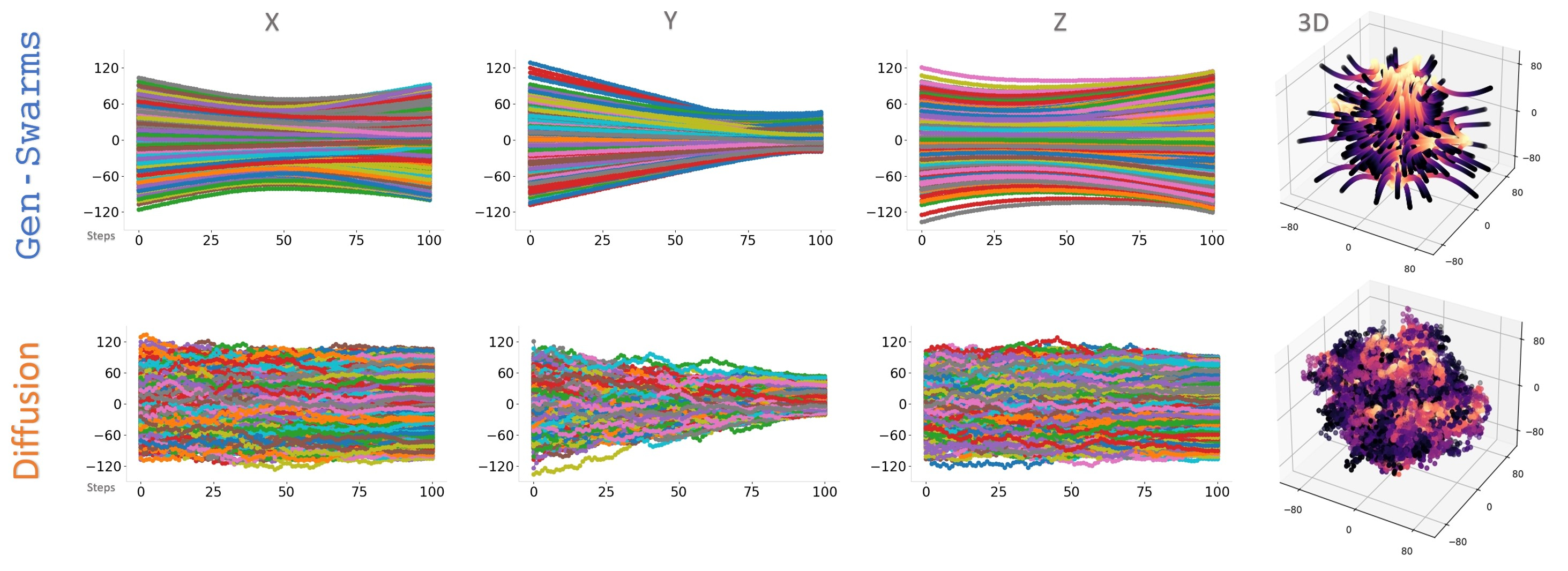}
    \caption{  
   \textbf{Trajectories.} Comparison of the generated trajectories of the drones using \texttt{Gen-Swarms} (top row) versus \emph{Diffusion} (bottom row). The first three plots show the evolution of each component ($x, y, z$ respectively) of the 3D trajectories of the drones, each drone showed with a different color. The fourth column displays the combined 3D trajectories.
}
    \label{fig:traj}
\end{figure}

\paragraph{Trajectories.} We compare the smoothness of the trajectories produced by \emph{Diffusion} and \texttt{Gen-Swarms} for each 3D component as the drones move to form a significant 3D shape, as illustrated in \cref{fig:traj}. The fourth column displays the evolution of all particles in 3D using a magma color scale (purple for the initial state and yellow for the final state), highlighting the notable qualitative differences between the trajectories generated by each method, particularly when considering their use in a drone show.

\begin{figure}[!tb]
    \centering
\includegraphics[width=\linewidth]{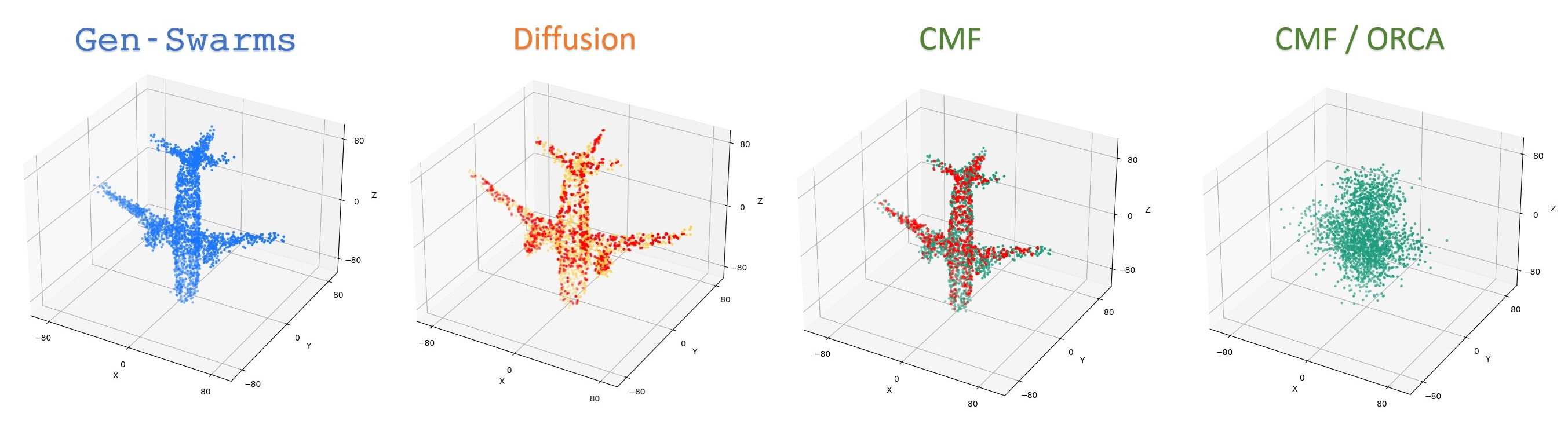}
    \caption{\textbf{Final shapes and collisions.} Final 3D point cloud generated by each method. Drones marked in red represent a violation of the security threshold, leading to a collision.}
    \label{fig:final_shapes_collisions}
\end{figure}

\paragraph{Final shapes and collisions.} We illustrate in \cref{fig:final_shapes_collisions} an example of a final 3D shape produced by each method in the \emph{Airplane} category. Drones highlighted in red would violate the minimum security distance ($\kappa = 2 \text{ m.}$ in the experiments), potentially leading to collisions, which are notably present in \emph{Diffusion} and \emph{CFM}. To ensure a fair comparison, we have selected the most similar shapes generated by each method.

\begin{figure}[!tb]
    \centering
\includegraphics[width=\linewidth]{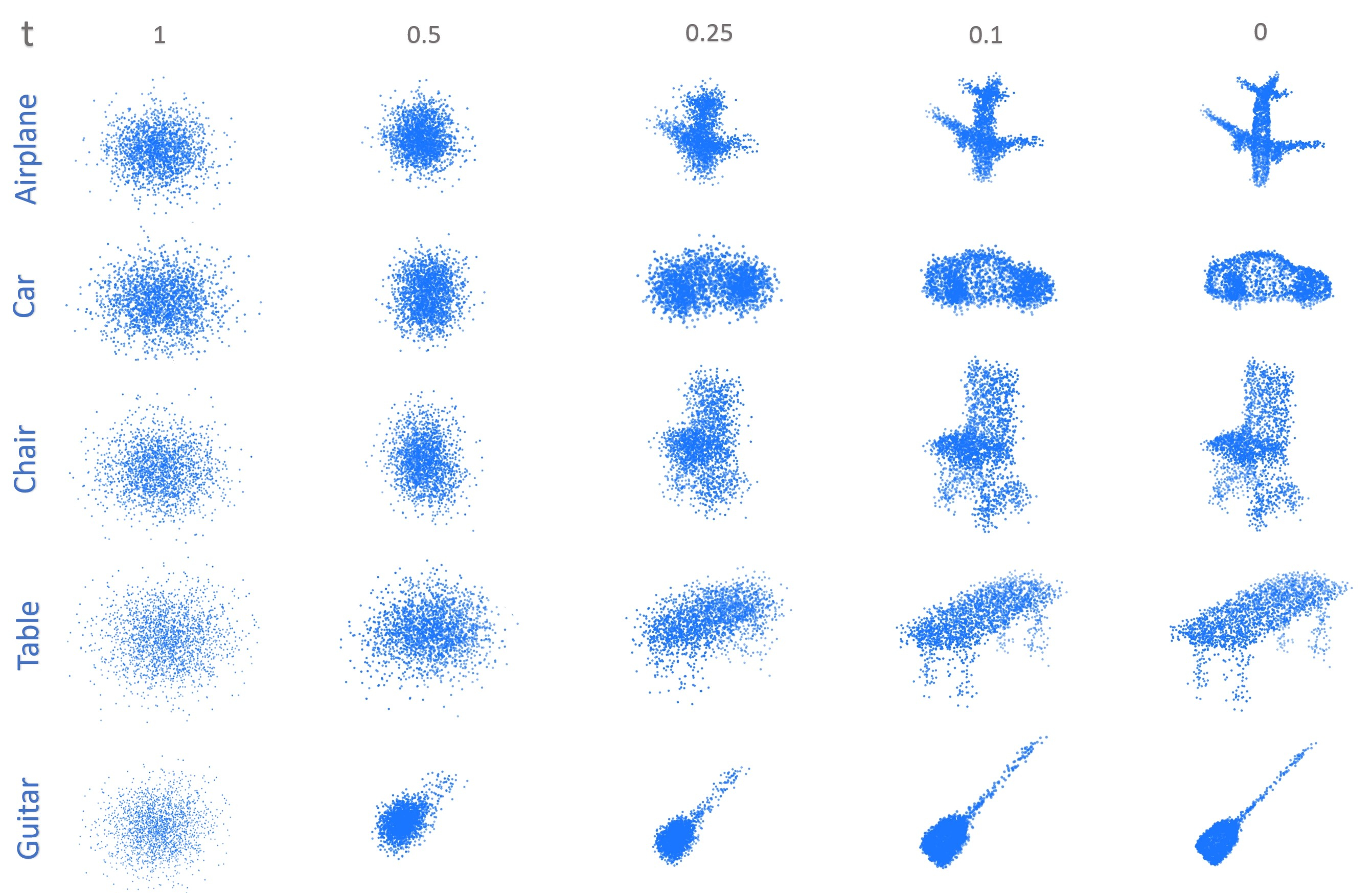}
    \caption{\textbf{Evolution of the 3D point cloud. }Point clouds at different stages of the \texttt{Gen-Swarms} sampling, showing the evolution towards the final shapes for five different categories: \textit{Airplane, Car, Chair, Table}, and \emph{Guitar}.}
    \label{fig:examples}
\end{figure}

\paragraph{Evolution of the 3D point cloud.} All the figures provided so far are for the category \textit{Airplane}. In \cref{fig:examples}, we present additional examples of 3D shapes for other categories (i.e., \textit{car, chair, table, guitar}) generated by \texttt{Gen-Swarms}. These examples also depict the evolution of the shapes through the algorithm's steps, starting from random noise.

\begin{figure}[!tb]
\includegraphics[width=\linewidth]{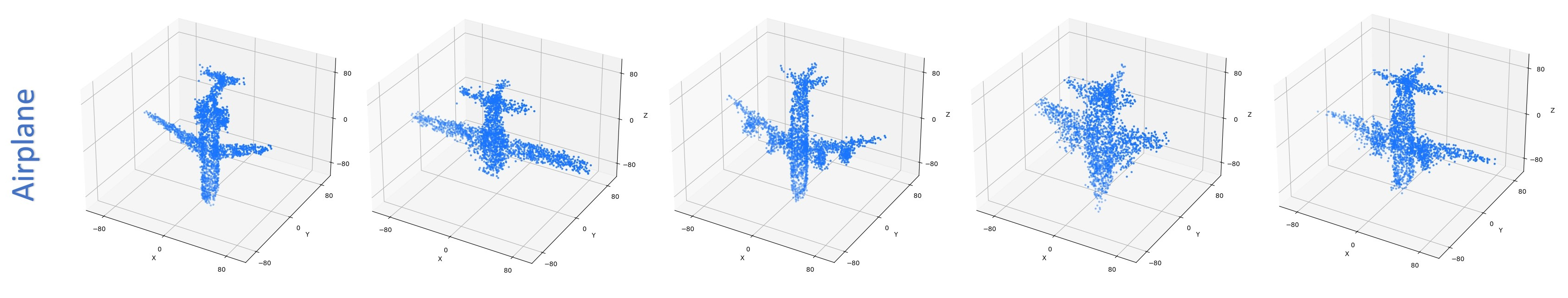}
    \caption{\textbf{Generative attributes of \texttt{Gen-Swarms}}: Different final shapes for the class \textit{Airplane} generated by \texttt{Gen-Swarms}, each exhibiting distinct attributes.}
    \label{fig:Airplanes}
\end{figure}

\paragraph{Generative attributes of }\texttt{Gen-Swarms}. In \cref{fig:Airplanes}, we display various final point clouds generated for the category \textit{Airplane}. These point clouds differ from one another in their characteristics, highlighting the generative capabilities of \texttt{Gen-Swarms}.

\section{Conclusions}
\label{sec:conclusions}
In this paper, we introduced \texttt{Gen-Swarms}, a novel framework that combines the strengths of deep generative models with reactive navigation algorithms for automated drone show creation. Our approach successfully integrates conditional flow matching algorithm in 3D point cloud generation framework to produce high-quality, dynamically guided drone performances.

Our experiments highlight that \texttt{Gen-Swarms} generates realistic and various 3D shapes based on text categories and coordinates smooth, collision-avoiding drone trajectories. The flow matching approach has proven particularly effective in this context, offering feasible trajectories and high-quality shape representations, which significantly enhance the overall performance of drone show generation. Specifically, our method outperforms the baselines in terms of trajectory smoothness, collisions, and energy consumption, while remaining competitive in terms of quality 3D shape reconstruction.  This advancement not only enhances the visual impact of drone shows but also paves the way for automatic generation of complex aerial displays.

Future work will focus on employing more sophisticated architectures to integrate color data alongside the positional information of each point, as well as developing an open-vocabulary approach to handle any text-based query as input. Additionally, an important challenge for applying our current method to real drone shows would be to effectively manage the transition between different shapes without needing to initialize from a random point cloud.

Overall, \texttt{Gen-Swarms} represents a significant step forward in automating drone show creation and holds exciting potential for future advancements in this field.

\section*{Acknowledgments}
This work was supported by a DGA scholarship and project T45\_23R;
MCIN/AEI/ ERDF/NextGenerationEU/PRTR projects PID2021-125514NB-I00, PID2021-125209OB-I00 and TED2021-129410B-I00, and by ONR research grants N62909-24-1-2081, and N00014-23-1-2354.
%
%
\bibliographystyle{splncs04}
\bibliography{main}

\end{document}